\documentclass[letterpaper, 10pt, conference]{ieeeconf}
\IEEEoverridecommandlockouts
\usepackage{flushend}
\usepackage{marginnote}
\usepackage{graphics}
\usepackage[pdftex]{graphicx}
\DeclareGraphicsExtensions{.pdf,.png,.jpg}
\usepackage[font={small}]{caption}
\usepackage{subcaption}
\usepackage[rightcaption]{sidecap}
\usepackage{pbox}
\usepackage{mathtools}
\usepackage{amsmath, amssymb, amscd}
\usepackage{ wasysym } %
\usepackage{amsfonts}
\usepackage{mathptmx} %
\usepackage{gensymb} 
\DeclareMathAlphabet{\mathcal}{OMS}{lmsy}{m}{n}
\DeclareSymbolFont{largesymbols}{OMX}{cmex}{m}{n}
\usepackage{textcomp} %
\usepackage[ruled,vlined,linesnumbered]{algorithm2e} 
\usepackage{algorithmicx, algpseudocode} 
\usepackage{array} %
\usepackage{tabularx}
\usepackage{multirow}
\usepackage{multicol}
\usepackage{booktabs}
\usepackage[T1]{fontenc} 
\usepackage[utf8]{inputenc}
\usepackage[english]{babel} %
\usepackage{units}
\usepackage{bm}
\usepackage{times} %
\usepackage{xspace}
\usepackage{flushend}%
\usepackage{csquotes}
\usepackage{makeidx}
\usepackage{soul} %
\usepackage{subfiles} %
\let\labelindent\relax
\usepackage{enumitem}

\usepackage[protrusion=true,expansion=true]{microtype}
\usepackage[yyyymmdd]{datetime}

\date{\protect\formatdate{1}{1}{2001}}

\usepackage{url}
\makeatletter
\g@addto@macro{\UrlBreaks}{\UrlOrds}
\makeatother
\usepackage{color}
\usepackage[usenames,dvipsnames,table]{xcolor}
\usepackage[pdfborder={0 0 0.5}]{hyperref}
\hypersetup{
    colorlinks=true,
    linkcolor=black,
    citecolor=black,
    filecolor=cyan,
    urlcolor=black
}

\newcommand{\argmax}{\operatornamewithlimits{arg\ max}}

\usepackage{amsmath}
\usepackage{balance}
\usepackage{caption}
\numberwithin{equation}{section}

\begin{document}

\title{\LARGE \bf
Deformable Elasto-Plastic Object Shaping using an \\ Elastic Hand and Model-Based Reinforcement Learning

}

\author{Carolyn Matl, Ruzena Bajcsy
\thanks{All authors are affiliated with the Department of Electrical Engineering and Computer Science, University of California, Berkeley, CA, USA;
\newline
\texttt{{\{carolyn.matl, bajcsy\}@eecs.berkeley.edu}}
}}

\maketitle

\begin{abstract}
Deformable solid objects such as clay or dough are prevalent in industrial and home environments. However, robotic manipulation of such objects has largely remained unexplored in literature due to the high complexity involved in representing and modeling their deformation. This work addresses the problem of shaping elasto-plastic dough by proposing to use a novel elastic end-effector to roll dough in a reinforcement learning framework. The transition model for the end-effector-to-dough interactions is learned from one hour of robot exploration, and doughs of different hydration levels are rolled out into varying lengths. Experimental results are encouraging, with the proposed framework accomplishing the task of rolling out dough into a specified length with 60\% fewer actions than a heuristic method. Furthermore, we show that estimating stiffness using the soft end-effector can be used to effectively initialize models, improving robot performance by approximately 40\% over incorrect model initialization.
\end{abstract}

\IEEEpeerreviewmaketitle

\section{Introduction}

Robotic manipulation of deformable objects has applications in various domains, including robotic surgery, home robotic solutions, and automated food preparation. A ubiquitous challenge that persists throughout this area of robotics research is that robotic interactions with deformable material are highly complex, often nonlinear, and difficult to model. These complexities impose additional challenges of tractability in action selection for manipulating the deformable object. The particular task this work focuses on is shape-servoing elasto-plastic solid (3-dimensional) deformable objects. Specifically, the goal is to find a sequence of rolling actions to apply to a ball of dough to efficiently form it into a desired length. Rolling out dough into a log is a crucial part of many dough manipulation tasks, from dumpling making to bread shaping.

To address this task, we designed an elastic end-effector attachment for a parallel jaw gripper that can vary in stiffness via stretching. This design is a revision of the StRETcH sensor \cite{matl2021soft}, a variable stiffness tactile end-effector. Changes were incorporated to compensate for kinematic limits of the robot arm and to allow for deeper contacts with the dough. Soft robotic end-effectors have low intrinsic stiffness and mass, making them particularly useful for handling delicate objects such as coral reefs \cite{galloway2016soft} and wine glasses \cite{alspach2019soft}. However, they can be challenging to manufacture or require additional tethering, cabling, and hardware. In contrast, the proposed elastic end-effector attachment is easy to assemble and can be made from two elastic bands. Furthermore, with the aid of a depth camera rigidly mounted above the work surface, it can be used to estimate the stiffness of objects with an active palpation technique developed in prior work \cite{matl2021soft}. 

\begin{figure}[t]
	\includegraphics[width=\linewidth]{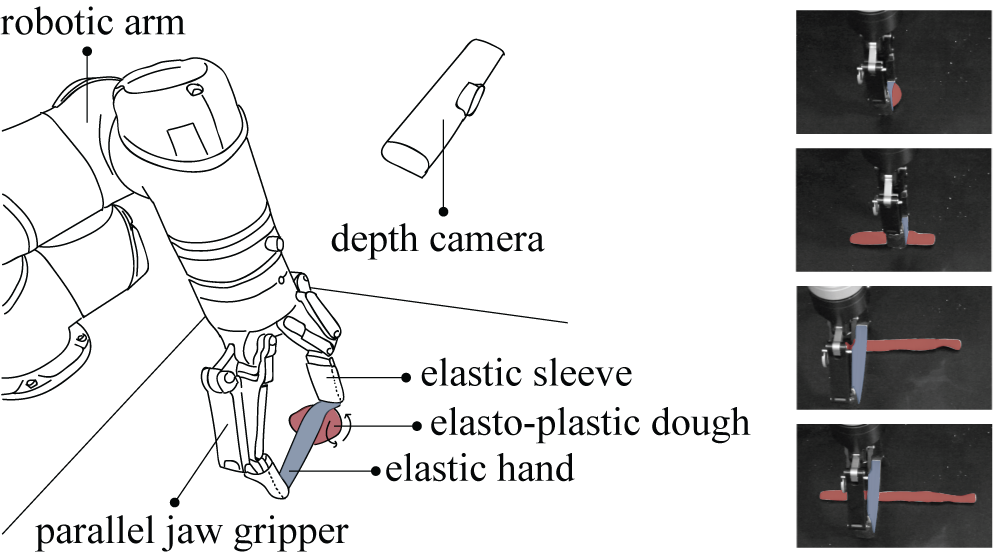}
	\centering
	\caption{(left): Diagram of the robotic system. The elastic hand is used to roll the elasto-plastic dough back and forth. The gripper changes the hand's stiffness via stretching, which modulates the force imposed by the hand onto the dough. The overhead depth camera is used to track the progression of the dough's shape. (right): Four frames taken from a rollout of a dough ball into a 10 inch log. Image is colorized to match the diagram on the left. }
	\label{fig:figure1}
\end{figure}

Manipulation of clay or dough is typically non-prehensile, i.e., it involves pushing or moving the object against a surface \cite{cherubini2020model}. In this work, we design a simple action space to roll out a piece of dough against a horizontal flat surface and track the dough's shape progression with a low-dimensional feature space. These simplifications help condense the complexity of the task, making model-based reinforcement learning a tractable solution. Dynamics models are learned for specific dough through random exploration, and experiments demonstrate that model-based reinforcement learning is an effective way to complete the given task of dough rolling. Furthermore, we show that stiffness estimation can be used to match the correct dough model for more efficient rolling. To the authors' best knowledge, this is the first work to propose a solution using a soft robotic end-effector to shape-servo a deformable elasto-plastic object.

The main contributions of this paper are:
\begin{itemize}
    \item A novel elastic end-effector attachment designed for shaping solid deformable objects that is cost-effective and easy to integrate with a parallel-jaw gripper. The elastic end-effector is also used to estimate the stiffness of different dough, which improves rolling performance. 
    \item An algorithm leveraging model-based reinforcement learning and simple featurization and actions that enable tractable optimization for the real-time solid deformable object manipulation task.
    \item Robotic experiments that roll doughs of different hydration levels into varying lengths. These experiments serve to quantify and demonstrate the efficacy of the proposed hardware and software framework. 
\end{itemize}

\section{Related Work}

\subsection{Deformable Object Manipulation}

Commonplace in many domains like the home environment, deformable objects are categorized as linear, planar, and solid objects \cite{sanchez2018robotic}. Efforts in robotic manipulation of deformable objects have largely focused on manipulating linear objects such as rope \cite{yan2020learning, sundaresan2020learning} and cables \cite{zhu2018dual}, as well as planar objects like paper \cite{balkcom2008robotic}, cloth \cite{miller2012geometric, matas2018sim} and gauze \cite{thananjeyan2017multilateral}. This particular work is interested in studying the manipulation of solid deformable objects like dough, which remains the least researched category within robotic deformable object manipulation \cite{sanchez2018robotic}.  

The complexities involved in representing a solid deformable object's shape as well as modeling its interactions with a robot and the environment contribute to the difficulty of this area of research. Thus, there has been significant effort in designing effective ways to parameterize deformable objects. For instance, a variety of approaches have been proposed for tracking the contour or shape of a deformable object during manipulation, ranging from continuum mechanics and simulation-based methods  \cite{petit2017tracking, giiler2015estimating} to neural network-based methods for contour segmentation and deformation tracking \cite{cretu2010deformable, cretu2011soft, staffa2015segmentation}. However, due to the nature of this work's defined task and its objective, simple geometric features are demonstrated to be a sufficient parameterization of the dough. 

In terms of designing control or action selection policies to effectively manipulate deformable solid objects, prior work has utilized both model-based \cite{higashimori2010active, yoshimoto2011active} and model-free adaptive methods \cite{navarro2016automatic, cherubini2020model}. While simulation-based methods have been shown to be quite useful in the space of linear and planar object manipulation \cite{sundaresan2020learning, ganapathi2020learning}, it is more computationally expensive to collect large amounts of data for physically-realistic simulations of dough. Learning from demonstrations has therefore been cited to be the most popular method for training a robot to shape deformable objects \cite{nadon2018multi}, accomplishing tasks such as shaping sand \cite{cherubini2020model} and pizza dough \cite{figueroa2016learning}. Similar to the work in \cite{higashimori2010active, yoshimoto2011active}, this work uses a model-based method, where material-relevant parameters are estimated and then used to guide the control strategy. However, the model does not explicitly represent the deformations of the object but rather captures its transitions resulting from robot interactions. Because this work is model-based, it requires significantly less data than model-free methods.

This work is unique in that it is the first work to propose using a soft robotic end-effector as the tool for shape-servoing dough. Furthermore, intentional design of simple feature and action spaces allows for real-time model-based reinforcement learning to solve the desired task.

\subsection{Soft Robotic End-effectors}

A soft robotic end-effector was designed specifically for the task of rolling and shaping dough. Composed entirely of an elastic band, the end-effector's variable stiffness allows it to be sensitive to highly deformable objects while capable of imposing enough force to exert work on the object in order to change its shape. Soft robotic end-effectors have been used for grasping objects as delicate as coral reef \cite{galloway2016soft}, maneuvering fragile wine glasses \cite{alspach2019soft}, and supporting and interacting with humans \cite{huang2020high}. The ability to conform to the environment and controllably vary in stiffness is crucial when handling soft, elasto-plastic objects like dough, which will permanently deform under large forces or flatten from rigid end-effectors. 

Soft robotic end-effectors have also been coupled with tactile sensing. Capable of providing high-resolution images of contact area, soft tactile sensors have been used to detect tumors in tissue \cite{ward2018tactip} as well as estimate the hardness of fruits \cite{yuan2017shape}. In this work, the proposed end-effector is a variation of the Soft to Resistive Elastic Tactile Hand (StRETcH) \cite{matl2021soft}, which is a variable stiffness tactile end-effector made of elastic material. In \cite{matl2021soft}, StRETcH was used to measure both the stiffness of different objects as well as their geometry under contact. Preliminary experiments showed promising results in using variable stiffness to roll a block of Play-Doh into a sphere via a heuristic circular motion. We hope to expand on that concept by pairing the new iteration of StRETcH with an action-selection algorithm to roll a ball of Play-Doh into logs of different lengths.

\subsection{Model-Based Reinforcement Learning}

The proposed action-selection algorithm for this work is rooted in model-based reinforcement learning. Reinforcement learning (RL) has been used to manipulate linear and planar objects using both model-free \cite{lin2020softgym, matas2018sim, jangir2020dynamic} and model-based approaches \cite{han2017model, ebert2018visual}.  Much of this prior work relies on simulation in order to collect training data, as physical experiments are expensive. However, simulated representations of three-dimensional deformable objects like clay or dough are complex or physically inaccurate and thus have not been substantially explored. Model-free RL has been shown to be effective at learning control policies directly from high-dimensional input, although these methods require a considerable amount of data, while model-based RL is much more sample efficient at the cost of scaling poorly with high-dimensional tasks \cite{lyumbb}. This work largely takes inspiration from the model-based RL framework proposed in Nagabandi et. al. \cite{nagabandi2018neural}, which highlights the sample efficiency achieved with a model-based approach. Furthermore, the parameterization of observations and simplification of the action space ensures that this task is low-dimensional and therefore optimization is tractable. Data collection for training the transition model of the nonlinear end-effector-to-dough interactions is simulation-free and only uses approximately one hour of real robot time per dough.  Experiments demonstrate that model-based RL efficiently accomplishes the task of dough rolling, even for poorly initialized models. 

\section{Problem Statement}

\subsection{Definitions and Objective}

The objective of this work is to design a policy to select a sequence of rolling actions to apply to a ball of dough, using the elastic end-effector, to form it into a desired length. More specifically, at time step $t$, the dough will be in a state $\mathbf{s}_t\in\mathcal{S}$. In reality, $\mathbf{s}_t$ will represent a partial observation of the true state of the dough. Using the elastic end-effector, the robot can apply a rolling action $\mathbf{a}_t\in \mathcal{A}$ to the dough. Upon execution of the action, the dough will transition to a new state $\mathbf{s}_{t+1}$.  This transition is assumed to be captured by an underlying dynamics or transition function $f: \mathcal{S}\times\mathcal{A}\rightarrow \mathcal{S}$. 

Since the goal is to find a sequence of actions that will form the dough into the desired shape in the least amount of steps, the problem can be naturally formulated in a reinforcement learning (RL) framework. In RL, a reward $r_t = r(\mathbf{s}_t, \mathbf{a}_t)$ is received at each time step $t$, and the robot or agent is tasked with taking the action that will maximize the cumulative reward over future time steps. We design the reward function to directly reflect task progression, i.e., the shape of the dough at time $t$ ($\mathbf{s}_t$) relative to the goal state $\mathbf{s}_g$. Thus, an $H$-length sequence of future actions can be selected by evaluating the objective function at each time step $t$:
\begin{equation}
    (\mathbf{a}_t, \dots, \mathbf{a}_{t+H-1}) = \argmax_{\mathbf{a}_t, \dots, \mathbf{a}_{t+H-1}} \sum_{n=0}^{H-1} \gamma^n r(\mathbf{s}_{t+n}, \mathbf{a}_{t+n})
    \label{eq:objectivefunction}
\end{equation}
where $\gamma \in [0,1]$ is a discount factor that places more weight on more immediate rewards. 

\subsection{Assumptions}

The interaction between an elastic membrane and an elasto-plastic object is complex and difficult to model, but the following assumptions allow for a tractable, real-time solution. First, when the robot is not in contact with the dough, we assume that the dough's state is in equilibrium and unchanging. This allows for the simplification of transition dynamics, where each state observation $\mathbf{s}_t$ is taken at a time step $t$ when the robot is no longer in contact with the dough. The transition function $f$ models the change in state due to this applied action and thus is not time-dependent. Rather, the problem can now be considered in a discrete-time formulation, where samples are observed between what are assumed to be stationary interactions. Additional assumptions include that the dough is uniform, it is not sticky or softer than the elastic membrane when un-stretched, and there exists enough friction between the dough and the work surface so that the dough will roll instead of slide under applied forces from the end-effector. Furthermore, the goal states are assumed to be reachable from the initial spherical state of the dough. With these assumptions, explicitly modeling the complex interactions between the dough and the soft elastic end-effector can be circumvented. Instead, the transition function $f$ summarizes the bulk effects due to the applied actions at a sparse timescale that is sufficiently descriptive and computationally friendly. 

\section{Methods}

\begin{figure}[t]
	\includegraphics[width=\linewidth]{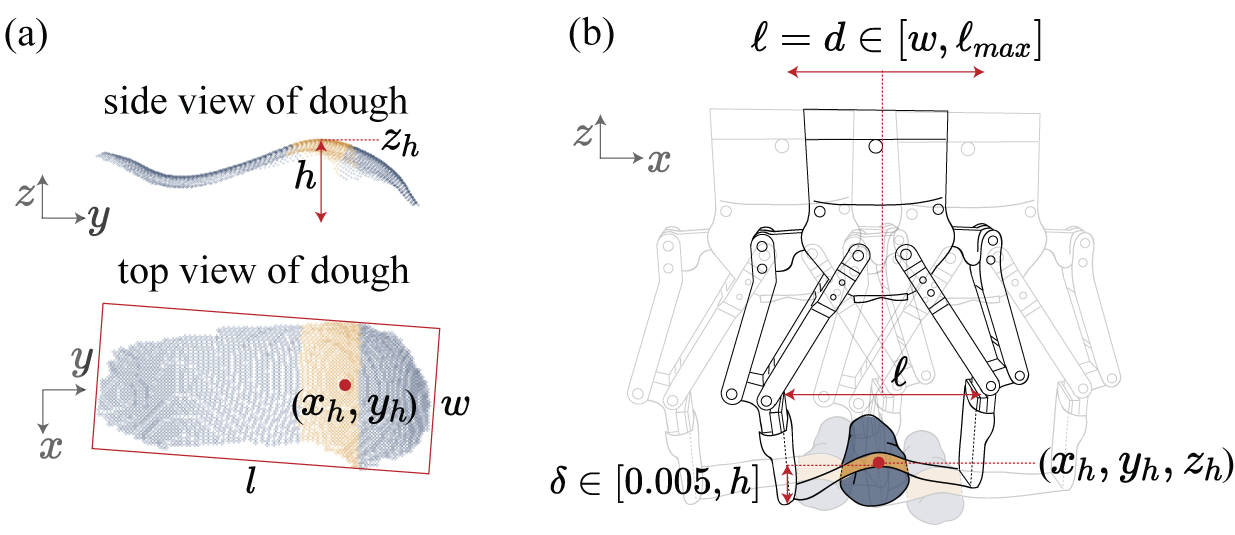}
	\centering
	\caption{(a): Featurization of a point cloud of dough. The state observation $[l,w,h]$ is extracted by finding the minimum bounding box enclosing this point cloud. The yellow region signifies where the robot will contact the dough in the next roll. (b): Each roll is state-dependent. The length of the elastic band $\ell$ and the distance of the roll $d$ are fixed to be equivalent and lower-bounded by $w$. The depth of the indentation $\delta$ is upper-bounded by $h$.}
	\label{fig:featurization}
\end{figure}

\subsection{Featurization of Dough}

Due to the objective's geometric nature, a low-dimensional feature vector was used to capture the shape of the dough. The chosen state space is defined as $S=\{l,w,h, x_h, y_h\}$, a five-dimensional space composed of the length, width, and height (in meters) of the minimum bounding box enclosing the dough, and the $x-y$ position of the highest point of the dough ($x_h, y_h, z_h$). The goal states are uniform cylinders of varying lengths, so this low-dimensional feature space is sufficient for the task. For more complicated objectives, e.g., forming a shape that cannot be parameterized by simple geometric features, a larger state space can be explored.

To extract these features from the original point cloud of the dough, a depth image of the workspace is sampled when the end-effector is not in contact with the dough. The depth image is projected into a 3D point cloud using camera intrinsics, and the point cloud is transformed into world coordinates, with the origin centered at the base of the robot. The workspace surface and the robot end-effector are filtered from the transformed point cloud, resulting in a point cloud of the surface of the dough. The highest point ($x_h, y_h, z_h$) is determined from the filtered point cloud. $h$ is calculated by subtracting the maximum height ($z_h$) of the point cloud from the height of the workspace surface, and $l$ and $w$ are calculated by finding the minimum-area bounding rectangle of the points, projected into the $x-y$ plane.  Feature extraction is illustrated for an example point cloud in Figure \ref{fig:featurization}a. 

\subsection{Simplified Action Space}

While we acknowledge the diversity of actions that can achieve the desired goal states, we greatly simplify the search problem by reducing the action space to five dimensions. To do this, we constrain most of the rolling motion applied at each step $t$. Furthermore, added heuristic constraints compress the effective search space to two dimensions. 

First, let a rolling motion $m(\ell, d, x, y, z)$ be defined by the length of the elastic band $\ell$ (i.e., the gripper throw), the distance of the roll $d$, and the position $(x,y,z)$ of the robot hand where the roll begins and ends. Each rolling motion is constrained to move in a back and forth motion along the direction of the $x$-axis, which should lie approximately perpendicular to the length of the dough formation. Thus, the entire roll begins with the robot end-effector positioned at $(x,y,z)$, moves to position $(x+\alpha*\frac{d}{2}, y, z)$, position $(x-\alpha*\frac{d}{2}, y, z)$, and ends at position $(x,y,z)$, where $\alpha$ is randomly assigned $1$ or $-1$ each time $t$ for symmetry. 

To further simplify the search space over actions $\mathcal{A} \in \{\ell, d, x, y, z\}$, a heuristic policy targets every roll to begin at the $x-y$ position of the highest point of the dough formation $(x_h, y_h, z_h)_t$. That is, the roll motion fixes $x=x_h$ and $y=y_h$. Roll distance $d$ is also constrained to equal the parallel jaw gripper throw $\ell$. These heuristic constraints keep the dough fairly even during rollouts and reduce the effective action space. Thus, the two remaining variables for the rolling motion $m$ are $\ell$ and $z$. We lower bound the range of $\ell$ at time $t$ by the width of the dough $w$ at that time step. Varying $\ell$ subsequently varies the stiffness of the elastic band, which consequently changes the force applied to the dough during rolling contact. Finally, the vertical depth at which the elastic band presses into the dough can be calculated by $\delta = z_h - z$. We impose the following bounds to the depth $\delta \in [0.005, h]$ to guarantee contact with the dough for all rolls while ensuring that the robot does not violate physical boundaries. The effective two-dimensional action search space used to parameterize the rolling motion thus consists of the length of the elastic band $\ell$ and the pressing depth $\delta = z_h - z$. These are summarized in the diagram in Figure \ref{fig:featurization}b. Note that all roll motions are executed at the same speed.

\subsection{Random Exploration}

Random interactions with the dough are used for two purposes -- to collect data to learn a dynamics function $f$ and to use as comparison against the proposed action-selection algorithms. At time step $t$, a random action $\mathbf{a}_t=(\ell_t, \delta_t)$ is uniformly sampled from a state-dependent distribution. Specifically, the length of the elastic band $\ell_t$ is sampled from a uniform distribution bounded by $[w_t, \ell_{max}]$, where $w_t$ is the current width of the dough and $\ell_{max}$ is the maximum gripper throw. The pressing depth $\delta_t$ is sampled from a uniform distribution bounded by $[0.005, h_t]$. Once the random action $a_t$ is sampled, a rolling motion $m(\ell_t, \ell_t, x_h, y_h, z_h-\delta_t)$ is executed, i.e., the gripper is commanded to stretch the elastic band to length $\ell_t$, the distance of the roll is $\ell_t$, and the start and end position of the roll is defined by $(x_h, y_h, z_h-\delta_t)$.

Random exploration thus consists of a sequence of actions sampled from the updating state-dependent bounds for an indeterminate amount of time for off-policy learning or until the goal state is reached. We determine that the goal state is reached if the Euclidean distance from the current state $\mathbf{s}_t$ to the desired state $\mathbf{s}_g$ is less than a pre-defined $\epsilon$.

\subsection{Heuristic Rolling}

For the proposed heuristic rolling method, each roll is still applied at the position of the highest point of the dough formation $(x_h, y_h)$, but a fixed policy is used for action selection. $\ell_t$ is fixed at $\frac{4}{5}\ell_{max}$ for high force output, and $\delta_t$ is fixed at $z_h - (z_g/2)$. At first, $\delta_t$ was fixed at $z_h - z_g$, or the difference in height of the current state to the goal state. However, this proved to be an inefficient heuristic as the dough state progressed toward the goal, since $\delta_t$ would gradually decrease and subsequently the imposed force would decrease as well. We therefore chose to fix $\delta_t$ at $z_h - (z_g/2)$ for greater force output throughout the rollout, although any choice of $\ell_t$ and $\delta_t$ is arbitrary without explicit knowledge of the hand-dough dynamics. These fixed actions are applied until the goal state is reached (the Euclidean distance from the current state to the desired state is less than $\epsilon$). 

\subsection{Dynamics Model}

Training data for the dynamics model is collected via random exploration, resulting in a series of trajectories $(\mathbf{s}_0, \mathbf{a}_0, \dots, \mathbf{s}_{N-1}, \mathbf{a}_{N-1}, \mathbf{s}_{N})$, each of which has $N$ transitions. Training set $\mathcal{D}_{off}$ consists of input state-action pairs $\{(\mathbf{s}_t, \mathbf{a}_t)\}$ with a corresponding output set of $\{\mathbf{s}_{t+1} - \mathbf{s}_t\}$. Note that only a subspace of the state space ($\{\ell, w, h\}$) and action space ($\{\ell, \delta\}$) is used to learn the dynamics model, since ($x_h, y_h$) in early experiments did not demonstrate significant effect on the subsequent transition, while the remaining action space was heuristically constrained. The dynamics model $\hat{f}_\theta(\mathbf{s}_t, \mathbf{a}_t)$, parameterized by weights $\theta$, can be learned via multiple machine learning techniques. Given the low-dimensional state and action spaces, we chose to represent $\hat{f}_\theta$ with a linear regression model with polynomial features of degree 2. 

\subsection{Rolling with Model-Based Reinforcement Learning}

In order to perform action-selection using model-based reinforcement learning, we largely take inspiration from recent advances in model-based RL robotics research \cite{williams2017information, nagabandi2018neural, lambert2019low, nagabandi2020deep}. These works use Model Predictive Control (MPC) to select the next action to execute within a Reinforcement Learning framework to iteratively improve the dynamics model. Model Predictive Control solves the objective defined by \ref{eq:objectivefunction} at each time step $t$, executes the first action in the optimal sequence $a_t$, and repeats this process at the next time step once state $s_{t+1}$ is observed. MPC assumes knowledge of system dynamics, so the executed action sequence may be suboptimal given an incorrect or poorly-initialized model. To compensate for these errors, the reinforcement learning framework iteratively refines the dynamics model by refitting the weights of $\hat{f}_\theta$ every $T$ steps, using both the original data $\mathcal{D}_{off}$ along with the on-policy data collected during execution $\mathcal{D}_{on}$. 
\subsubsection{Iterative Random Shooting with Refinement (Cross Entropy Method)}
MPC for nonlinear systems is an active area of research, and there are several methods that can be used to optimize the objective function in Equation \ref{eq:objectivefunction}. One such method that is considered for this work is iterative random shooting with refinement, which is a cross-entropy method (CEM) approach \cite{botev2013cross}.  Random-sampling shooting samples $K$ candidate action sequences of length $H$ for a planning horizon of $H$ steps and selects the action sequence corresponding to the highest cumulative reward ($\sum_{n=0}^{H-1} \gamma^n r(\mathbf{s}_{t+n}, \mathbf{a}_{t+n})$), where future states $\mathbf{s}_{t+n}$ for $n \in [1, H)$ are predicted via the trained dynamics model $\hat{f}_\theta(\mathbf{s}_{t+n-1}, \mathbf{a}_{t+n-1})$. The CEM approach improves random-sampling by iteratively refining the mean and variance of the sampling distribution for $M$ iterations, taking the mean of the last distribution as the optimal action sequence. In other words, for each iteration, random-sampling shooting generates $K$ candidate action sequences by sampling:
$$a_t^k \sim \mathcal{N}(\mu_t^m, \Sigma_t^m), \forall k \in K, t \in [0,H), m \in [0,M)$$
After $K$ sequences are sampled, the top $N$ sequences ($A_{top}$) corresponding to the highest cumulative rewards are used to refine the mean and variance of the next iteration:
$$ \mu_t^{m+1} = \beta\text{mean}(A_{top}) + (1-\beta)\mu_t^m, \forall t\in[0,H) $$
$$ \Sigma_t^{m+1} = \beta\text{var}(A_{top}) + (1-\beta)\Sigma_t^m, \forall t\in[0,H) $$
where $\beta$ is used to smooth the updates.
After $M$ iterations, $\mu_t^M \forall t \in [0,H)$ is chosen as the optimal action sequence. 
\subsubsection{Numerical Optimization with Powell's Method}

An alternative gradient-free approach is to maximize the objective in Equation \ref{eq:objectivefunction} using a numerical technique such as Powell's conjugate direction method. Given the low-dimensional action and state space, for a short-term planning horizon, this remains a tractable solution. We use Scipy's minimize function implementation of Powell's method to solve for the optimal sequence $(\mathbf{a}_t, \dots, \mathbf{a}_{t+H-1})$, given Equation \ref{eq:objectivefunction} and the transition function $\hat{f}_\theta(\mathbf{s}_t, \mathbf{a}_t)$ to estimate future states.

\subsection{Model Matching via Stiffness Estimation}
\label{sec:stiffness_estimation}

\begin{figure}[t]
	\includegraphics[width=0.6\linewidth]{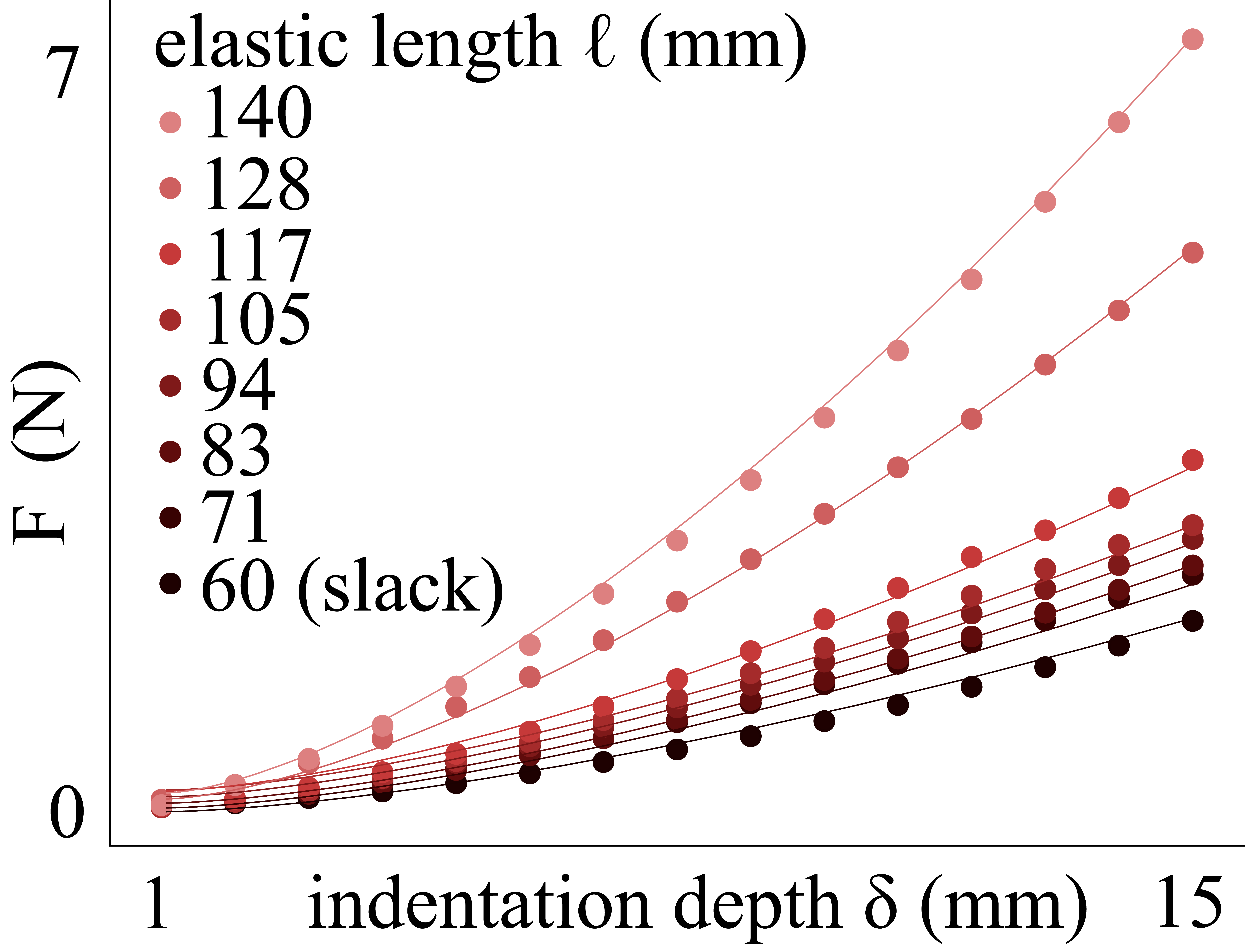}
	\centering
	\caption{Indentation depth $\delta$ vs. load force $F$ for varying stretch states
for a 30mm diameter hemispherical rigid indenter}
	\label{fig:force}
\end{figure}

In prior work, we demonstrate how a proxy for the stiffness of deformable objects can be estimated by palpating the objects from above and mapping the indentation depth and stretch state of the elastic hand to a predicted output force $F$ \cite{matl2021soft}. That is, we fit a model $F=h(\ell, \delta)$, and a one-dimensional measure for stiffness is calculated by dividing $F$ by the object's $z$ deflection. As discussed in \cite{matl2021soft}, $h(\ell, \delta)$ is also dependent on contact geometry, but given that stiffness estimation is used only for initializing the model, and the dough is assumed to be spherical with a known diameter, this work does not consider the effect of contact geometry when estimating the imposed force. A reasonable estimate of force can be made by first measuring the varying force while modulating the indentation depth and stretch state of the elastic hand against a rigid, hemispherical probe (of equivalent diameter as the dough) attached to an ATI Axia80 EtherNet Force/Torque sensor. These force measurements are depicted in Figure \ref{fig:force} and the model $F=h(\ell,\delta)$ is fit to this data. As shown in Figure \ref{fig:force}, force output increases with increasing indentation depth and stretch. When estimating the stiffness of an object, the robot actively palpates the object until a deflection is observed by the overhead camera. At the point of deflection, the predicted force is divided by the amount of deflection, giving a proxy estimate of stiffness in $N/mm$. 

We can use these proxy estimates of stiffness to initialize dynamics models for unknown dough. The stiffness of a dough is dependent on its hydration level, or its water content, and stiffness directly affects the material properties and subsequently the nonlinear dynamics of the dough. In Section \ref{sec:doughconsistency}, we discuss how we design a soft dough, a stiff dough, and a medium-stiff dough by varying their hydration levels. We train a separate model for the dry and hydrated dough. However, to estimate the dynamics of an unknown dough, e.g. the medium-stiffness dough, we can linearly interpolate the predictions from the two given models by weighting their predictions based on stiffness. That is, let the stiffness and model of the dry and hydrated dough be defined as $\sigma_{dry}$, $\hat{f}_{dry}(\cdot)$, $\sigma_{hydrated}$, and $\hat{f}_{hydrated}(\cdot)$. If the estimated stiffness of the unknown dough is $\sigma_{unknown}$, then the corresponding dynamics model for the new dough is:
\begin{equation}
    \mathbf{s}_{t+1} = (1-\beta)*\hat{f}_{hydrated}(\mathbf{s}_t, \mathbf{a}_t) + \beta*\hat{f}_{dry}(\mathbf{s}_t, \mathbf{a}_t) 
    \label{eq:blenddynamics}
\end{equation}
where $\beta=\frac{\sigma_{unknown}-\sigma_{hydrated}}{\sigma_{dry}-\sigma_{hydrated}}$ and $\sigma_{unknown}$ is bounded by $[\sigma_{hydrated},\sigma_{dry}]$. Stiffness is likely not simply linearly related to the resulting dynamics of the dough, but later experimental sections show that initializing the dynamics model by this method still improves performance over incorrectly initializing it to either $\hat{f}_{hydrated}(\cdot)$ or $\hat{f}_{dry}(\cdot)$.

\section{Experimental Setup}
\subsection{Elastic Hand Design}
The design of the elastic hand is a variation of StRETcH \cite{matl2021soft}, a soft tactile hand attached to a parallel jaw gripper that varies in stiffness via opening and closing the gripper. In StRETcH, tactile images are observed via an overhead camera mounted rigidly to the robot arm. However, the configuration of the elastic membrane along the length of the robot fingers was kinematically limiting for this task. Thus, for this work, the StRETcH sensor was redesigned with the elastic surface rotated by 90 degrees (see Figures \ref{fig:figure1} and \ref{fig:featurization}).  Furthermore, the only materials used to construct this new end-effector attachment were two elastic bands. The stretchier band was used as the soft contact interface for the dough. The stiffer elastic material was sewn into two separate loops, which were sewn to each end of the stretchier band. The stiff loops acted as sleeves that were securely attached to the robot fingers via friction fit, positioning the softer band between the fingers. This attachment was cost-effective to make and required no additional hardware or cabling. 

\subsection{Physical Setup}

The elastic end-effector attachment was secured via friction fit to a Robotiq 2-Finger Adaptive Robot Gripper, which was attached to a Universal Robot UR5 arm. In contrast to prior work \cite{matl2021soft}, in which the camera was mounted rigidly to the robotic arm, this work fixes the Intel RealSense Depth Camera SR305 at a stationary position, looking down at a slight angle onto the workspace surface. The reachable workspace surface was approximately 1ft$\times$1ft. 

\subsection{Dough Consistency}
\label{sec:doughconsistency}

Play-Doh was used as the elasto-plastic material for all experiments. Some of the following experiments explored rolling dough of varying stiffnesses. To create three different stiffnesses from the same Play-Doh, each 50 gram portion of Play-Doh was hydrated at a different hydration level. To hydrate the dough, different amounts of water were kneaded into the dough until the water was completely absorbed. The more water incorporated, the softer the dough became. 

Using the method developed in prior work \cite{matl2021soft} and described in Section \ref{sec:stiffness_estimation}, the robot measured the stiffness of 50 grams of dough when hydrated with $1$, $2$, $3$, $4$, and $5$ grams of water. The approximate stiffnesses were $1.23$, $0.97$, $0.85$, $0.67$, and $0.49$ N/mm, respectively. Thus, the three doughs used for experiments in Sections \ref{sec:experiment2} and \ref{sec:experiment3} were hydrated with approximately $1$, $3$, and $5$ grams of water. Beyond $6$ grams, the dough became too sticky to shape using the elastic membrane. The rest of this paper refers to the doughs as:

\begin{table}[ht]
\centering
\resizebox{0.35\textwidth}{!}{
  \begin{tabular}[t]{cccc}
    \hline 
     dough & A & B & C \\
     \hline 
     water (g) & 5 & 3 & 1 \\
     stiffness (N/mm) & 0.49 & 0.85 & 1.23 \\
     \hline 
  \end{tabular}}
  \caption{Hydration levels and proxy stiffness values of doughs A, B, and C.}
\label{tab:hydrations}
\end{table}

\subsection{Data Collection}
We collected a total of two hours of training data, with one hour each for dough A and C. Each initialization began with the dough hand-rolled into a roughly spherical shape. The dataset for dough A consisted of four initializations, with 150 random actions sampled for each. For dough C, the training trajectories consisted of six initializations, with 100 random actions sampled for each. Random exploration was restarted with a new initialization once rate of progress (dough-lengthening) appeared to diminish substantially, which accounts for the difference in initializations between dough A and C.  Each dough dataset had a total of 600 actions and transitions, and a separate model was trained for each dataset. 

\section{Results}
\subsection{Comparison of Rolling Algorithms}
\begin{figure*}[t]
	\includegraphics[width=\linewidth]{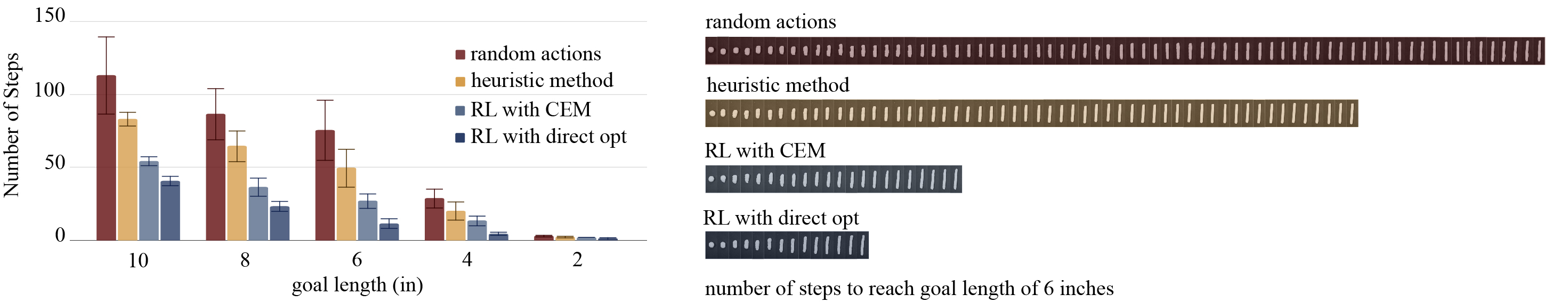}
	\centering
	\caption{(left) Number of steps to roll a ball of dough into the goal length for four different action-selection methods. 6 trials are represented per bar and error bars reflect a 95\% confidence interval. (right) An example of the dough state progression at each time step for all four methods.}
	\label{fig:algs_comparison}
\end{figure*}

We begin by comparing the four proposed action-selection methods: random sampling, a heuristic method with a fixed action policy, RL with random shooting and CEM, and RL using numerical optimization. Both RL algorithms use MPC for action selection and iteratively refine the dynamics model as the robot experiences new data. For both RL methods, the planning horizon ($H$) and model iteration horizon ($T$) were both chosen to be $10$ steps, and the CEM method additionally used $K=100$, $M=3$, $\beta=0.5$ as hyperparameters. New data $\mathcal{D}_{on}$ were weighted $10\times$ more than $\mathcal{D}_{off}$ when re-fitting $\hat{f}_\theta(\cdot)$. For this experiment, all rolling was performed on dough A, and the dynamics models were initialized to $\hat{f}_A$. For each method, six independent rollouts were run for five different goal lengths of $2$, $4$, $6$, $8$, and $10$ inches, and the number of steps until the goal was reached was recorded. 

Figure \ref{fig:algs_comparison} summarizes the results of the experiment. The heuristic method was on average $27.6\pm6.1\%$ more efficient than purely random sampling, and both RL methods were more efficient than the heuristic method ($33.4\pm18\%$ and $63.1\pm18.3\%$, respectively). Numerical optimization was more efficient than RL with CEM across all goals, so the following experiments used numerical optimization. 

\subsection{Comparison of Fixed versus Iterative Models}
\label{sec:experiment2}
\begin{figure}[t]
	\includegraphics[width=0.9\linewidth]{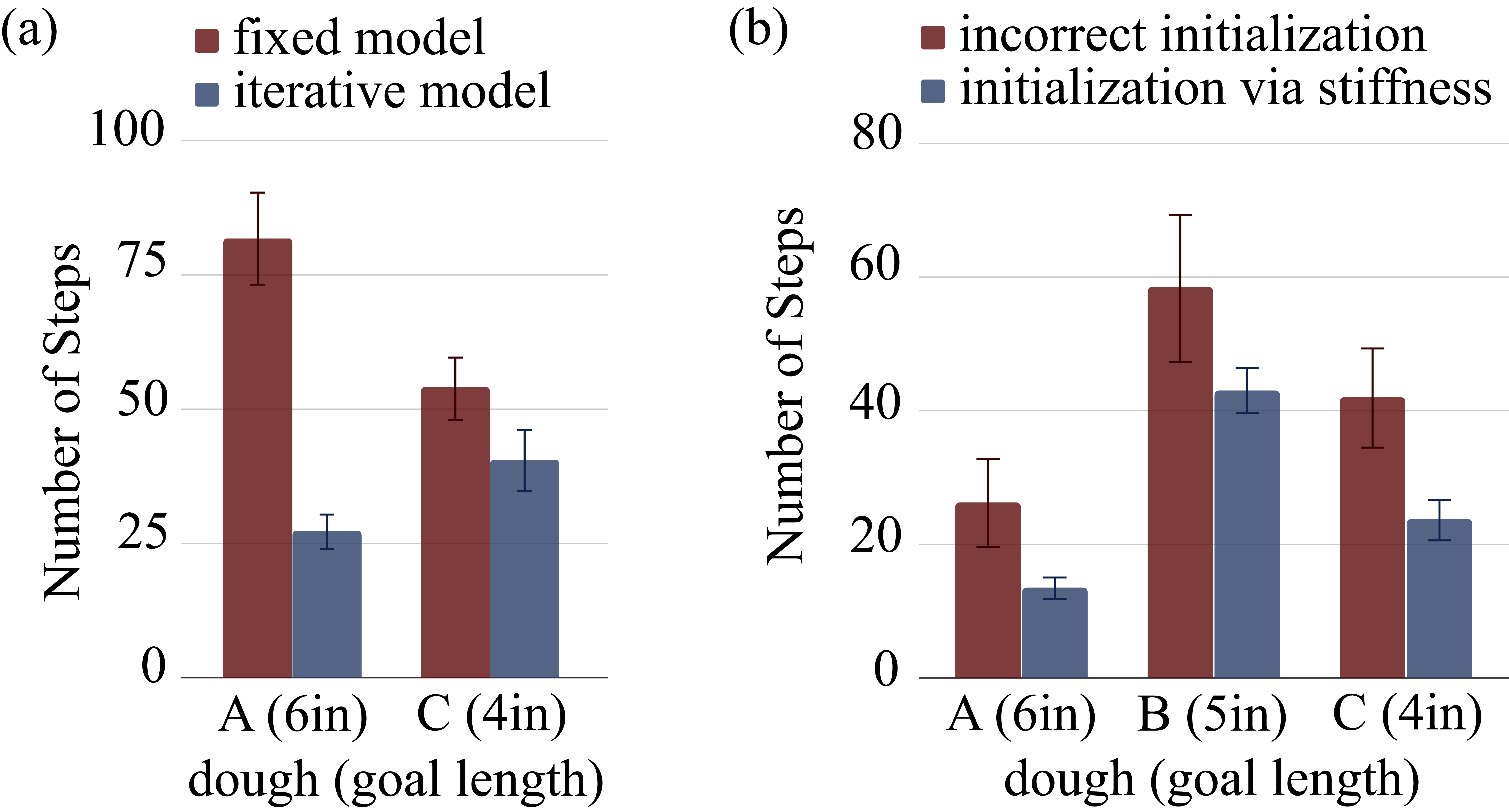}
	\centering
	\caption{(a): Comparison of using a fixed vs. iterative model for Model Predictive Control. The iterative model is refined every $T=5$ steps from new data collected at run-time. In both scenarios, the initial model is incorrect (dough A is initialized with $\hat{f}_C$ and vice versa). The goal states for doughs A and C are $6in$ and $4in$ in length, respectively. 15 trials are represented per bar. (b): Comparison of incorrectly initializing the model vs. initializing the model using the estimated stiffness of the dough. Dough A is incorrectly initialized with $\hat{f}_C$, Dough C is incorrectly initialized with $\hat{f}_A$, and Dough B is initialized by either $\hat{f}_A$ or $\hat{f}_C$ with equal probability. The goal states for Doughs A, B, and C are $6in$, $5in$, and $4in$ in length, respectively.  10 trials are represented per bar.}
	\label{fig:fixedvsiterative_stiffnessmodelmatching}
\end{figure}

In the first experiment, the dynamics model was initialized correctly for the corresponding dough, and model-based RL methods achieved the goal state efficiently. In this experiment, we tested the robustness of the RL framework to iteratively correct a poorly-initialized or incorrect dynamics model. To do so, we compared using MPC to roll out dough using a fixed dynamics model versus an iteratively improved model using RL. Furthermore, for both cases, we incorrectly initialized the dynamics model -- dough A trials were initialized with $\hat{f}_C$, and similarly dough C trials were initialized with $\hat{f}_A$. This experiment aimed to test the self-correcting ability of the RL framework. The goal states of dough A and C were set to 6in and 4in in length, respectively. This difference was to compensate for the fact that dough C was stiffer and thus took longer to roll out than dough A. For this experiment, numerical optimization directly solved Equation \ref{eq:objectivefunction} with $H=5$ as the planning horizon at each step and $T=5$ steps were taken before each model iteration. As in the previous experiment, new data $\mathcal{D}_{on}$ were weighted $10\times$ more than $\mathcal{D}_{off}$ when re-fitting $\hat{f}_\theta(\cdot)$. For both fixed and iterative models, 15 rollouts were run per dough. 

As shown in Figure \ref{fig:fixedvsiterative_stiffnessmodelmatching}(a), the iterative model outperformed the fixed dynamics model using MPC. Using RL to iteratively refine the dynamics model reduced the number of steps to reach the goal by, on average, $45.8\%$. Furthermore, it was originally expected that MPC initialized by dough C would select more aggressive (high force output) actions to compensate for the higher stiffness and thus the fixed model would outperform the iterative model for dough A. However, as dough A's state got closer to $6in$, we observed actions by the robot that were less and less effective at inducing change in the dough's state. This was because the datapoints of dough C were very sparse in that goal range due to the nature of the stiffer dough and limited training time. However, by allowing model iteration via the RL framework, the model was quickly updated to better reflect the dough's true dynamics and the goal was reached in a short amount of time steps. 

Figure \ref{fig:action_comparisons} illustrates the average actions selected throughout the task, where the $x$-axis is the percent of total steps taken before the goal was reached. Comparing the fixed (red) and iterative (yellow) trends, one can see how the iterative model corrects throughout the progression of the task, although with considerably higher variance. For dough A, while the fixed model has the robot repeatedly performing a shallow roll with low $\ell$, the iterative model pulls the action space towards higher values for $\ell$. For dough C, the iterative model applied either high $\ell$ and low $\delta$ actions or low $\ell$ and maximum $\delta$ actions, resulting in high variance between the different rollouts. 

\subsection{Stiffness Estimation for Model Matching}
\label{sec:experiment3}

\begin{figure}[t]
	\includegraphics[width=\linewidth]{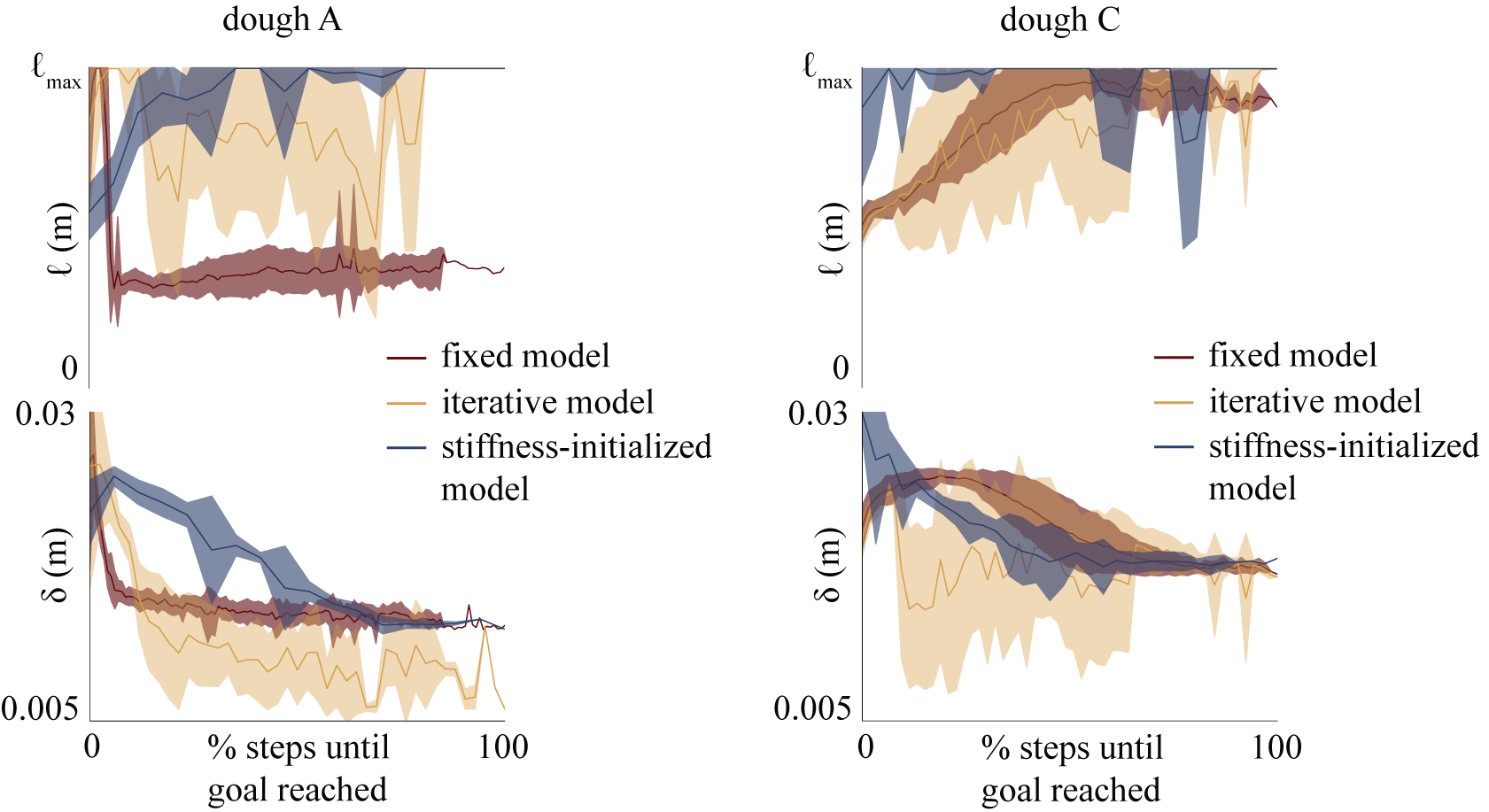}
	\centering
	\caption{(left): Comparison of actions chosen throughout the progression of the task for dough A, with red, yellow, and blue corresponding to the fixed incorrect dynamics model of $\hat{f}_C$, iterative dynamics model initialized at $\hat{f}_C$, and stiffness-initialized model, respectively. The top graph is the gripper throw (length of the elastic band, $\ell$) and the bottom graph is rolling depth $\delta$. (right): Comparison of actions for dough C, where the incorrect initialized dynamics model is $\hat{f}_A$. }
	\label{fig:action_comparisons}
\end{figure}

It is likely that robots will not have models for every dough they will need to manipulate. We have seen that the RL framework can compensate for an incorrect or poorly-trained model by iteratively refining the dynamics model. However, we hypothesized that estimating stiffness could improve model initialization and therefore execute the task more efficiently than initializing the dynamics model to a random dough. This was a reasonable proposition, since stiffness reflects material properties relevant to dough dynamics. 

To test this hypothesis, we compared using model-based RL (using numerical optimization) with incorrect initialization versus initialization via stiffness estimation. For each initialization method, 10 rollouts for doughs A, B, and C were run, with goal states of $6$, $5$, and $4$in, respectively. For incorrect initialization, dough A was initialized with $\hat{f}_C$, and dough C was initialized with $\hat{f}_A$. For dough B, for which there was no learned model, 5 of its rollouts were initialized with $\hat{f}_A$, and the other 5 were initialized with $\hat{f}_C$. For the initialization method that incorporated estimated stiffness, the dynamics model was initialized by Equation \ref{eq:blenddynamics}. This experiment used the same hyperparameters as in Section \ref{sec:experiment2}.

As shown in Figure \ref{fig:fixedvsiterative_stiffnessmodelmatching}(b), initializing the models based on stiffness was significantly more efficient than initializing the model incorrectly. On average, initializing the model via stiffness estimations improved performance by $39.6\%$. Dough B took more steps to reach its goal than doughs A and C, which may suggest that the interpolation method can be optimized to better reflect the relationship between stiffness and dough dynamics (i.e., this relationship is nonlinear). 

Figure \ref{fig:action_comparisons} compares the incorrectly initialized (yellow) and the stiffness-initialized (blue) models. The actions selected throughout the rollouts have considerably lower variance when the model is initialized by stiffness. The policy that seems to work effectively for dough A is initially rolling with low $\ell$ and high $\delta$, and gradually increasing $\ell$ throughout the rollout.  For dough C, the selected action sequences tended to prefer high $\ell$ and $\delta$ for high force-output, which is consistent with dough C's higher stiffness. Because many of the actions tended to prefer higher stiffness of the end-effector, a future consideration is to explore higher-stiffness elastic bands, or bands with greater ranges of variable stiffness. 

\section{Conclusions}

To the best of the authors' knowledge, this is the first work that accomplishes deformable manipulation using a soft end-effector on elasto-plastic dough. The novel end-effector is not only cost-effective and easy to integrate with a parallel-jaw gripper, but it also can vary in stiffness, which allows it to effectively manipulate soft deformable objects like dough. A model-based RL framework is leveraged to roll dough into different lengths, where simple featurization and actions enable tractable, real-time optimization. The robot is capable of accomplishing the task after only one hour of random exploration, and the RL framework iteratively improves the transition model to compensate for incorrect or poorly-defined models. Real robot experiments demonstrate high efficacy of the proposed system, with RL action-selection exceeding the heuristic method by more than $60\%$ in efficiency.  Furthermore, stiffness estimation using the soft end-effector enables better model initialization, which consequently improves the robot's performance by approximately $40\%$. 

We hope the experimental results of this work can inspire more efforts in tackling deformable solid object manipulation. Although the interactions involved in deformable object manipulation are highly complex and difficult to model, task-specific design of the state and action spaces can enable the robotic task to be much more tractable. 
In future work, we aim to explore other low-dimensional parameterizations of dough, including neural network-based encoders that may facilitate in tasks requiring the manipulation of dough into more complex geometries and shapes. Furthermore, we plan to experimentally compare more model representations, including Gaussian Process Regression and Neural Network Dynamic Models. Ultimately, we plan to use our findings to manipulate real dough into complex shapes such as pretzels and challah bread, which all require rolling dough into logs.

{\footnotesize
\section*{Acknowledgment} The authors were supported in part by the National Science Foundation Graduate Research Fellowship. We thank Andrea Bajcsy, Anusha Nagabandi, and Matthew Matl for their insightful feedback and suggestions.
}

\end{document}